\documentclass{article}
\usepackage{spconf,amsmath,graphicx}

\usepackage{url}
\usepackage{tabularx}
\usepackage{booktabs}
\usepackage{multirow}
\usepackage{microtype}
\usepackage{subfigure}
\usepackage{dialogue}
\usepackage[utf8]{inputenc}
\usepackage{textcomp}

\newcolumntype{C}{>{\centering\arraybackslash}X}
\newcolumntype{L}{>{\raggedright\arraybackslash}X}

\usepackage{cite}

\newcolumntype{b}{X}
\newcolumntype{s}{>{\hsize=.2\hsize}X}
\newcolumntype{e}{>{\hsize=.4\hsize}X}
\newcolumntype{f}{>{\hsize=.5\hsize}X}
\usepackage{xcolor}

\definecolor{mycolor}{HTML}{FF6600}

\title{Cross-lingual topic prediction for speech using translations}
\name{Sameer Bansal$^{1}$, Herman Kamper$^{2}$, Adam Lopez$^{1}$, Sharon Goldwater$^1$}
\address{
  $^{1}$School of Informatics, University of Edinburgh, UK \\
  $^{2}$Dept.\ E\&E Engineering Stellenbosch University, South Africa \\
{\small \tt sameer.bansal@ed.ac.uk, \{sgwater, alopez\}@inf.ed.ac.uk} \\ 
{\small \tt kamperh@sun.ac.za}
\\}

\begin{document}
%\ninept
%
\maketitle
\begin{abstract}
Given a large amount of unannotated speech in a low-resource language, can we classify the speech utterances by topic?
We consider this question in the setting where a small amount of speech in the low-resource language is paired with text translations in a high-resource language.
We develop an effective cross-lingual topic classifier by training on just 20 hours of translated speech,
using a recent model for direct speech-to-text translation. While the translations are poor, they are still good enough to correctly classify the topic of 1-minute speech segments over 70\% of the time—a 20\% improvement over a majority-class baseline. Such a system could be useful for humanitarian applications like crisis response, where incoming speech in a foreign low-resource language must be quickly assessed for further action.
\end{abstract}

\section{Introduction}
\label{sec:intro}
Quickly making sense of large amounts of linguistic data is an important application of language technology. For example, after the 2011 Japanese tsunami, natural language processing was used to quickly filter social media streams for messages about the safety of individuals, and to populate a person finder database \cite{neubig2011safety}. Japanese text is high-resource, but there are many cases where it would be useful to make sense of \emph{speech} in \emph{low-resource} languages. For example, in Uganda, as in many parts of the world, the primary source of news is local radio stations, which is broadcast in many languages. A pilot study from the United Nations Global Pulse Lab identified these radio stations as a potentially useful source of information about a variety of urgent topics related to refugees, small-scale disasters, disease outbreaks, and healthcare \cite{quinn2017using,menon+etal_interspeech19}. With many radio broadcasts coming in simultaneously, even simple classification of speech for known topics would be helpful to decision-makers working on humanitarian projects.

Speech classification systems have traditionally used automatic speech recognition (ASR) systems to first convert speech to text,
which is then used as input to a classifier.
However, this pipelined approach is impractical for unwritten languages, spoken by millions of people around the world.
Although transcriptions cannot be obtained in these settings, translations could provide a viable alternative supervision source~\cite{bird-EtAl:2014:Coling,blachon2016parallel,adda2016breaking,besacier2006towards}.
Recent research has shown that it is possible to train direct {\bf S}peech-to-text {\bf T}ranslation (ST) systems from speech paired only with translations~\cite{berard+etal_nipsworkshop16,weiss2017sequence,bansal2017towards}. 
Since no transcription is required, this
is useful in very low-resource settings.
However, in realistic low-resource settings where only a few hours of training data is available, these end-to-end ST systems produce poor translations~\cite{bansal+pretraining+arxiv+2018}. 
But it has long been recognized that there are good uses for bad translations~\cite{church1993good}.
Could classifying the original speech be another one
of these use cases?

\begin{figure}[t]
  \centering
  \includegraphics[width=0.85\linewidth]{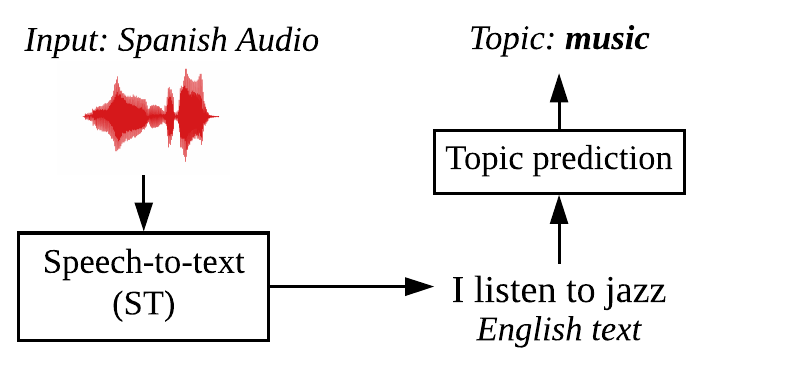}
  \vspace*{-12.5pt}
  \caption{Spanish speech is translated to English text, and a classifier then predicts its topic.}
  \label{fig:method}
\end{figure}

We answer this question affirmatively: we first use ST to translate speech to text, which we then classify by topic using supervised models (Figure~\ref{fig:method}). 
Although our ultimate goal is to work with truly low-resource languages, available datasets of this type are still too small to thoroughly evaluate and analyse.
We therefore test our method on a corpus of conversational Spanish speech paired with English text translations that has been widely used in ST research\cite{weiss2017sequence,bansal2018interspeech}, enabling us to put our results in context.
Using an ST model trained on 20 hours of Spanish-English data, we predict topics correctly 71\% of the time, and we outperform the majority class baseline with less than 10 hours of training data.
These promising results are the first we know of for this task, and open the door to future work on cross-lingual topic prediction from speech.

\section{Methods}

% \subsection{Speech-to-text}
\label{sub:method_st}
% \vspace{.1in}
\noindent {\bf Speech-to-text translation.} We use the method of Bansal et al.~\cite{bansal+pretraining+arxiv+2018} to train neural sequence-to-sequence Spanish-English ST models. As in that study, before training ST, we pre-train the models using English ASR data from the Switchboard Telephone speech corpus~\cite{LDC97S62}, which consists of around 300 hours of English speech and transcripts. In~\cite{bansal+pretraining+arxiv+2018} this was found to substantially improve translation quality when the training set for ST was only tens of hours.

\vspace{.1in}
\noindent {\bf Topic modeling and classification.}
\label{sub:method_topic_prediction}
To classify the translated documents, we first need a set of topic labels, which were not already available for our dataset. 
We therefore initially discover a set of topics from the target-language (English) training text using a topic model. To classify the translations of the test data, we choose the most probable topic according to the learned topic model.
To train our topic model, we use Nonnegative Matrix Factorization (NMF)~\cite{berry2007algorithms,arora+topic+nmf+2012}.%\footnote{
We also experimented with Latent Dirichlet Allocation~\cite{blei+lda+2003}, but manual inspection revealed that NMF produced better topics.

\section{Experimental Setup}
\label{sec:setup}

% \vspace{.1in}
\noindent {\bf Data.} \label{sub:telephone_calls_corpus} We use the Fisher Spanish speech corpus~\cite{LDC2010S01}, which consists of 819 phone calls, with an average duration of 12 minutes, giving a total of 160 hours of data.
We discard the associated transcripts and pair the speech with English translations~\cite{post2013improved}. 
To simulate a low-resource scenario, we sampled 90 calls (20h) of data ({\em train20h}) to train both ST and topic models, reserving 450 calls (100h) to evaluate topic models ({\em eval100h}). We investigate %Our experiments required
ST models of varying quality, so we also trained models with decreasing amounts of data: {\em ST-10h}, {\em ST-5h}, and {\em ST-2.5h} are trained on 10, 5, and 2.5 hours of data, respectively, sampled from {\em train20h}.
To evaluate ST only, we use the designated Fisher test set, as in previous work.

\vspace{.1in}
\noindent {\bf Fine-grained topic analysis.} \label{sub:topics_setup} In the Fisher protocol, callers were prompted with one of 25 possible topics.
% \footnote{Topics are shown in Appendix~\ref{sec:appendix_topics_assigned}.} 
It would seem appealing to use the prompts as topic labels, but we observed that many conversations quickly departed from the initial prompt and meandered from topic to topic. For example, one call starts: ``Ok today's topic is marriage or we can talk about anything else
\ldots.'' Within minutes, the topic shifts to jobs: ``I'm working oh I do tattoos.''
To isolate different topics within a single call, we split each call into 1-minute long segments to use as `documents'.
This gives 1K training and 5.5K test segments, but leaves us with no human-annotated topic labels for them. 

Obtaining gold topic labels for our data would require substantial manual annotation, so we instead use the human translations from the 1K ({\em train20h}) training set utterances to train the NMF topic model with {\em scikit-learn}~\cite{scikitlearn}, and then use this model to infer topics on the evaluation set. These {\em silver} topics act as an oracle: they tell us what a topic model would infer if it had perfect translations. 
% NMF and model hyperparameters are described in Appendix~\ref{sec:appendix_nmf}.

% \paragraph{Evaluation.}
To evaluate our ST models, we  apply our ST model to test audio, and then predict topics from the translations using the NMF model trained on the human translations of the training data (Figure~\ref{fig:method}).
To report accuracy we compare the predicted labels and silver labels, i.e.,
we ask whether the topic inferred from our predicted translation (ST) agrees with one inferred from a gold translation (human).

\section{Results}
\label{sec:results}
\begin{figure}
\centering
\includegraphics[width=.45\textwidth]{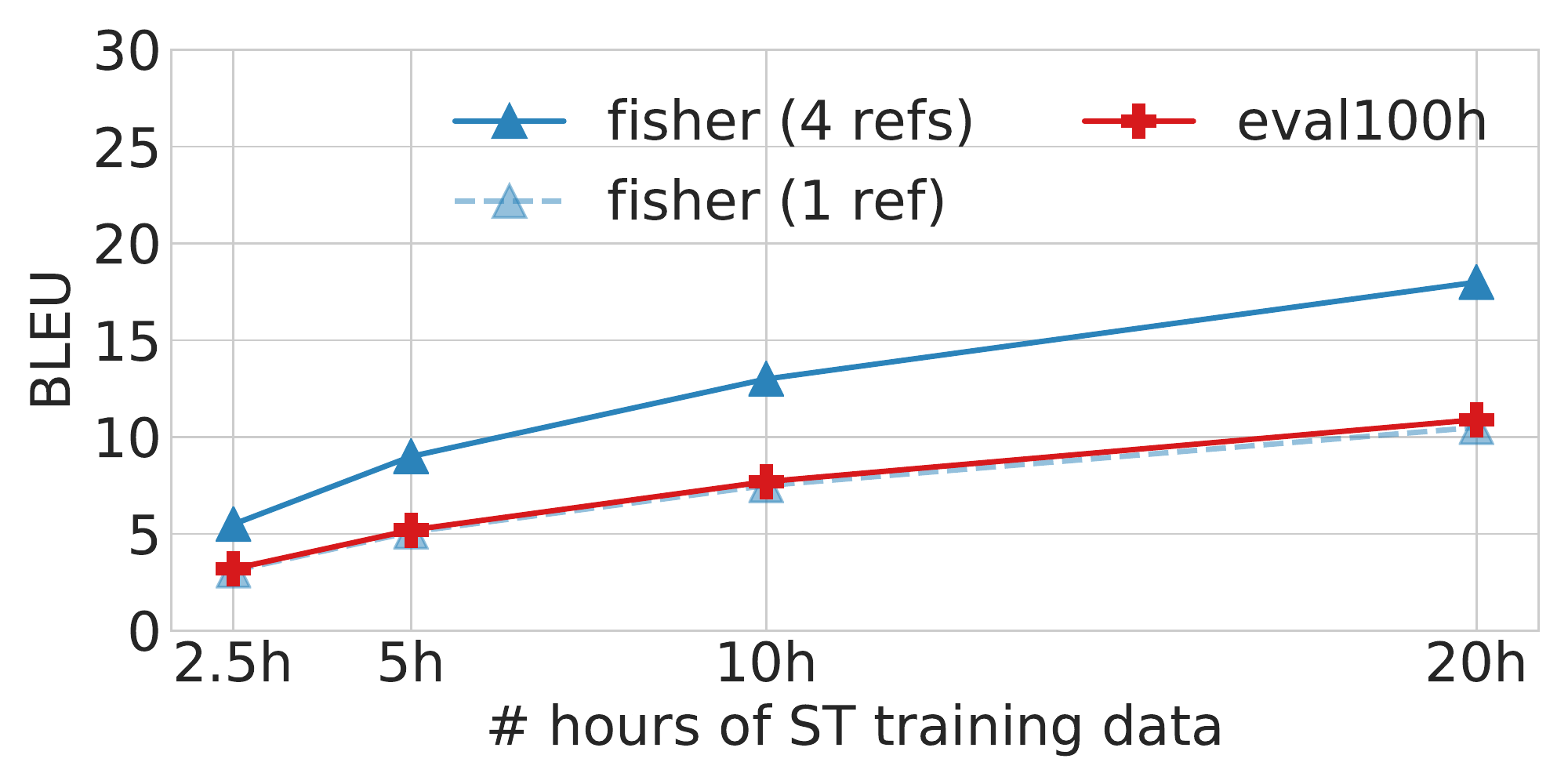}
\vspace*{-5pt}
\caption{BLEU scores for Spanish-English ST models computed on Fisher test set, using all 4 human references available, and using only 1 reference, and on {\em eval100h}, for which we have only 1 human reference.
}
\label{fig:st_bleu_1vs4refs}
\end{figure}

% \vspace{.1in}
\noindent {\bf Spanish-English ST.} \label{ssub:st_results} To put our topic modeling results in context, we first report ST results. Figure~\ref{fig:st_bleu_1vs4refs} plots the BLEU scores on the Fisher test set and on {\em eval100h} for Spanish-English ST models. The scores are very similar for both sets when computed using a single human reference; scores are 8 points higher on the Fisher test set if all 4 of its available references are used. The state-of-the-art BLEU score on the Fisher test set  is 47.3 (using 4 references), reported by~\cite{weiss2017sequence}, who trained an ST model on the entire 160 hours of data in the Fisher training corpus.
By contrast, our 20 hour model ({\em ST-20h}) achieves a BLEU score of 18.1.
Examining the translations (Table~\ref{tab:high_level_task}), we see that while they are mediocre, they contain words that might enable correct topic classification.

\begin{table}[t]
% \footnotesize
  \begin{center}
  \begin{tabularx}{0.98\linewidth}{r@{\hspace{1.5mm}}X}
  \toprule
    % &  & \multicolumn{2}{c}{\bf Expected output} \\
    {\bf audio} & yo eh oigo la música en inglés o americana \\
     {\bf human} & i eh \underline{listen} to \underline{music} in english or american \\
     {\bf ST} & i eh \underline{listen} to the \underline{music} in english \\
     {\bf topic} & {\em music} \\
     \midrule
     {\bf audio} & soy católica pero no en realidad casi no voy a la iglesia \\
     {\bf human} & i am \underline{catholic} but actually i hardly go to \underline{church} \\
     {\bf ST} &  i'm \underline{catholics} but reality i don't go to the \underline{church} \\
     {\bf topic} & {\em religion} \\
  \bottomrule
  \end{tabularx}
  \end{center}
  \vspace*{-5pt}
  \caption{Examples of Spanish {\bf audio} %input
  shown %here
  as Spanish text. An {\bf ST} system translates the audio into English text, and we give the {\bf human} reference.
  Our task is to predict the {\bf topic} of discussion in the audio, which are potentially signaled by the underlined words.
  }
  \label{tab:high_level_task}
\end{table}

\vspace{.1in}
\noindent {\bf Topic modeling on training data.}
\label{ssub:topic_modeling_results}
Turning to our main task of classification, we first review the set of topics discovered from the human translations of {\em train20h} (Table~\ref{tab:topics}). 
We explored different numbers of topics, and chose 10 after reviewing the results.
% \footnote{Appendix~\ref{sec:appendix_topics_assigned}-Table~\ref{tab:topics_n25_train160h} shows the result of using 25 topics, the same as the number of prompts in the Fisher protocol.}
We assigned a name to each topic after manually reviewing the most informative terms; for topics with less coherent % sets of
informative terms, we include {\em misc} in their names.

\begin{table}[t]
\begin{center}
%   \small
  \begin{tabularx}{\linewidth}{ll}
    \toprule
    {\bf Topic} & {\bf Most informative terms} \\
    \midrule
    family-misc & married, kids, huh, love, three \\
    music & music, listen, dance, listening, hear \\
    intro-misc & hello, fine, name, hi, york \\
    religion & religion, god, religions, believe, bible \\
    movies-tv & movies, movie, watch, theater \\
    welfare & insurance, money, pay, expensive \\
    languages-misc & english, spanish, speak, learn \\
    tech-marketing & phone, cell, computer, call, number \\
    dating & internet, met, old, dating, someone \\
    politics & power, world, positive, china, agree \\
  \bottomrule
  \end{tabularx}
  \end{center}
  \vspace*{-5pt}
  \caption{Topics discovered using human translated text from {\em train20h}, with manually-assigned topic names.
  }
%   \vspace{-1em}}
  \label{tab:topics}
\end{table}

For evaluation, silver labels are obtained by applying this topic model to human translations on the test data.
We argued above that the silver labels are sensible for evaluation despite not always matching the assigned call topic prompts, since they indicate what an automatic topic classifier would predict given correct translations and they capture finer-grained changes in topic.
Table~\ref{tab:sample_silver_review} shows a few examples where the silver labels differ from the assigned call topic prompts. In the first example, the topic model was arguably incorrect, failing to %our topic model hasn't 
pick up the prompt {\em juries}, and instead focusing on the other words, predicting {\em intro-misc}. But in the other examples the topic model is reasonable, %in fact
correctly identifying the topic in the third example where the transcripts indicate that the annotation was wrong (specifying the topic prompt as {\em music}).
In general, the topic model classifies a large proportion of discussions as {\em intro-misc} (typically at the start of the call) and {\em family-misc} (often where the callers stray from their assigned topic). 

Our analysis also supports our observation that discussed topics stray from the prompted topic in most speech segments. For example, among segments in the 17 training data calls with the prompt {\em religion}, only 36\% have the silver label {\em religion}, and the most frequently assigned label is {\em family-misc} (46\%). 
% Further details are in Appendix~\ref{sec:appendix_topics_drift}.

\begin{table}[t]
% \footnotesize
  \begin{center}
  \begin{tabularx}{0.98\linewidth}{Xcc}
    \toprule
    {\bf human translation} & {\bf Assigned} & {\bf Silver} \\
    \midrule
    hello good afternoon have you ever been in a jury in a trial & juries & intro-misc  \\
    \midrule
    i also receive many letters of life insurance from banks & spam & welfare \\
    \midrule
    they tell us we have to talk about marriage & music & family-misc \\
    \bottomrule
  \end{tabularx}
  \end{center}
  \vspace*{-5pt}
  \caption{Example audio utterances from {\em eval100h}. We show a part of the human translation here. {\bf Assigned} is the topic assigned to speakers in the current call to prompt discussion. {\bf Silver} is topic inferred by feeding the human translation through the topic model.}
  \label{tab:sample_silver_review}
\end{table}

\vspace{.1in}
\noindent {\bf Topic classification on test data.}
\label{ssub:topic_prediction_results}
We have four ST model translations: {\em ST-2.5h, 5h, 10h, 20h} (in increasing order of quality).
We feed each each of the audio utterences in {\em eval100h} into the topic model from Table~\ref{tab:topics} to get the topic distribution and use the highest scoring topic as the predicted label.

\begin{figure}[t]
\centering
\includegraphics[width=.45\textwidth]{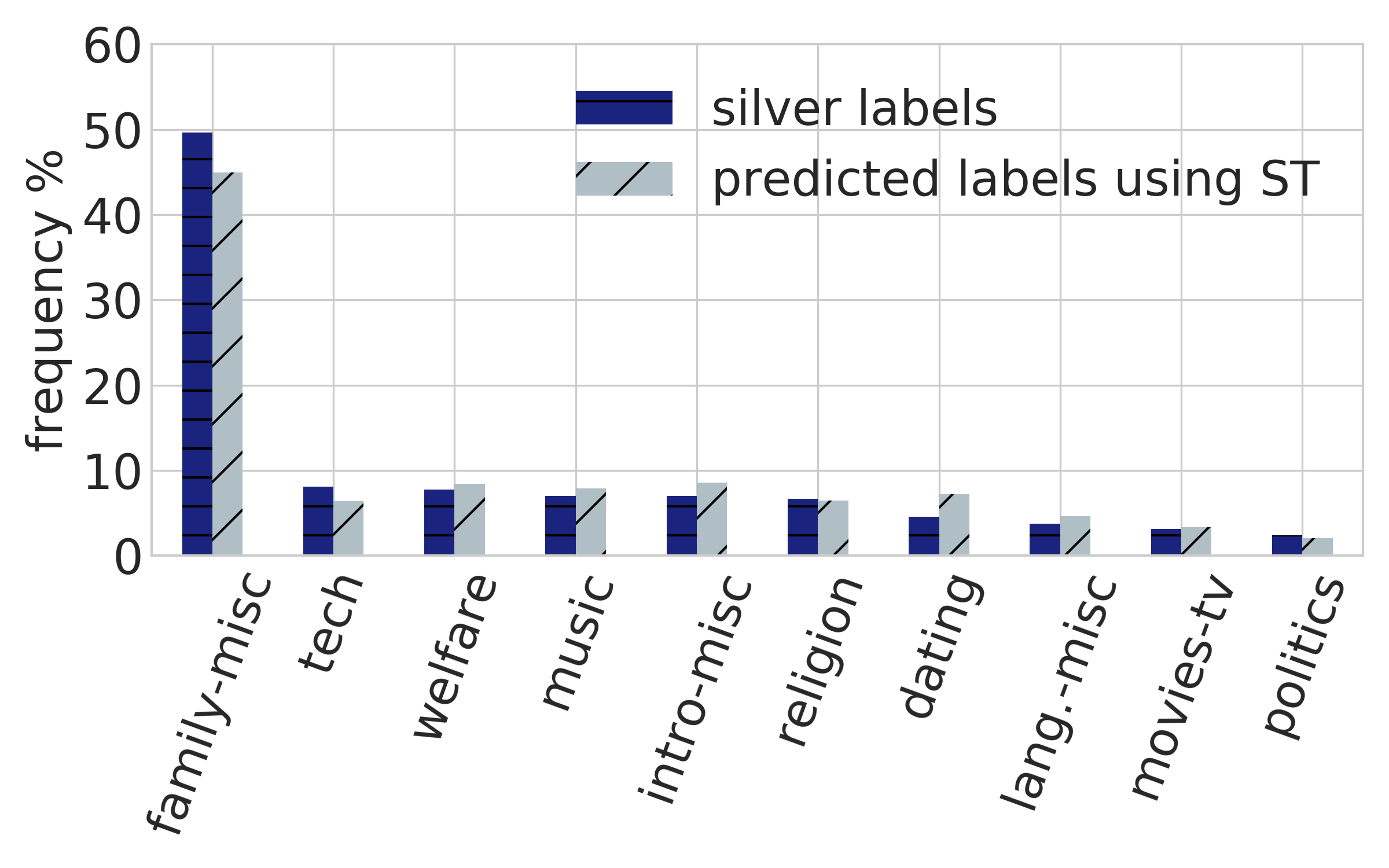}
  \vspace*{-5pt}
\caption{Distribution of topics predicted for the 5K audio utterances in {\em eval100h}. {\bf silver} labels are predicted using human translations. The {\bf ST} model has been trained on 20 hours of Spanish-English data.}
\label{fig:eval_topic_counts}
\end{figure}

Figure~\ref{fig:eval_topic_counts} compares the frequencies of the silver labels with the predictions from the {\em ST-20h} model.
The {\em family-misc} topic is predicted most often---almost 50\% of the time. This is reasonable since this topic includes words %normally
associated with small-talk.
Other topics such as {\em music}, {\em religion} and {\em welfare} also
occur with a high enough frequency to allow for a reasonable evaluation.

\begin{figure}[t]
\centering
\includegraphics[width=.45\textwidth]{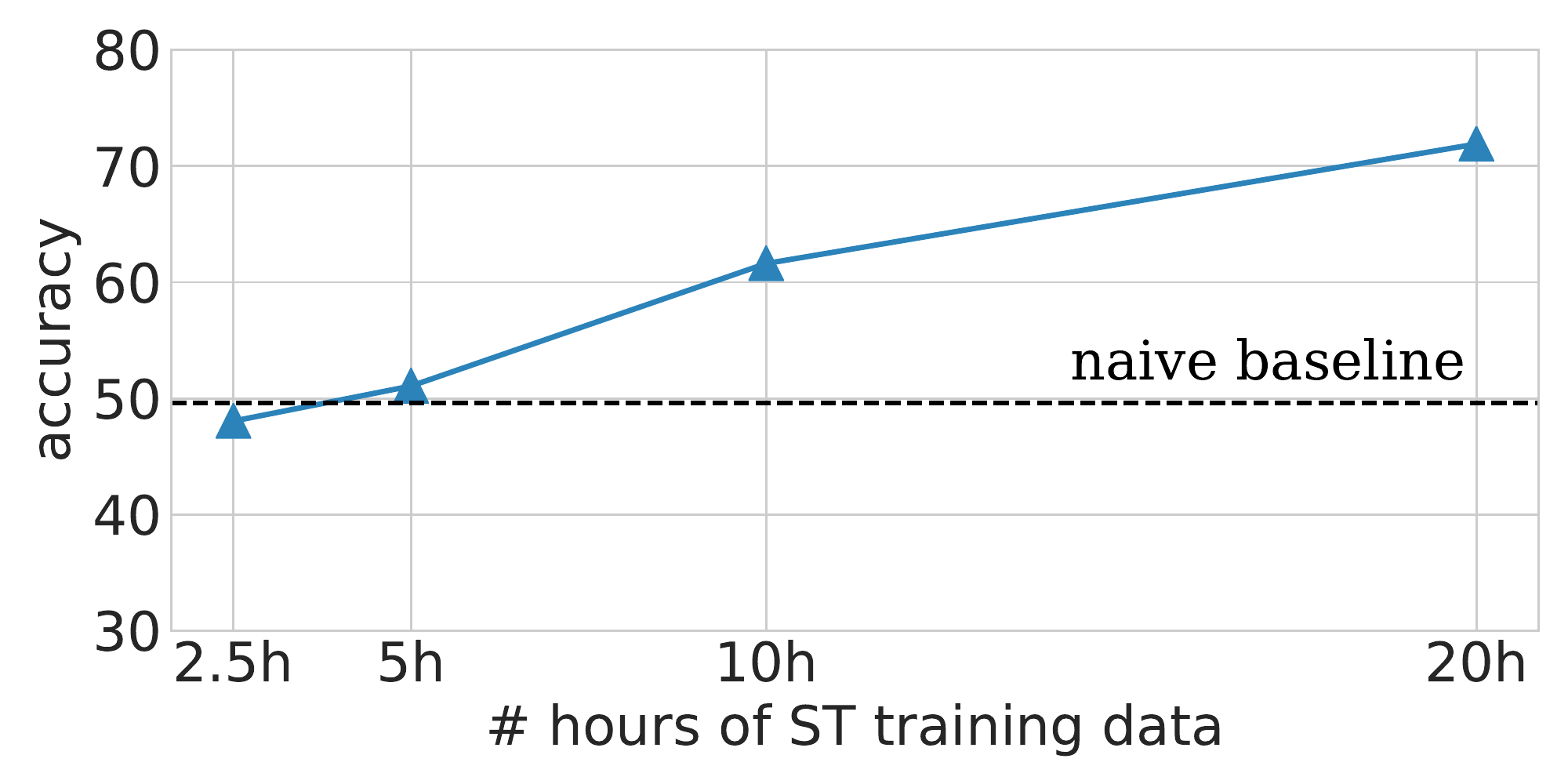}
  \vspace*{-5pt}
\caption{Accuracy of topic prediction using ST model output.
The {\bf naive baseline} is calculated using majority class prediction, which is the topic {\em family-misc}.}
\label{fig:eval_topics_accuracy}
\end{figure}

Figure~\ref{fig:eval_topics_accuracy} shows the accuracy for all ST models, treating the silver topic labels as the correct topics.
We use the {\em family-misc} topic as a majority class {\em naive baseline}, giving an accuracy of 49.6\%.
We observe that ST models trained on 10 hours or more of data outperform the {\em naive-baseline} by more than 10\% absolute, % points,
with {\em ST-20h} scoring 71.8\% and {\em ST-10h} scoring 61.6\%.
Those trained on less than 5 hours of data score close to or below that of the naive baseline: 51\% for {\em ST-5h} and 48\% for {\em ST-2.5h}.

\begin{figure}[t]
\centering
\includegraphics[width=.45\textwidth]{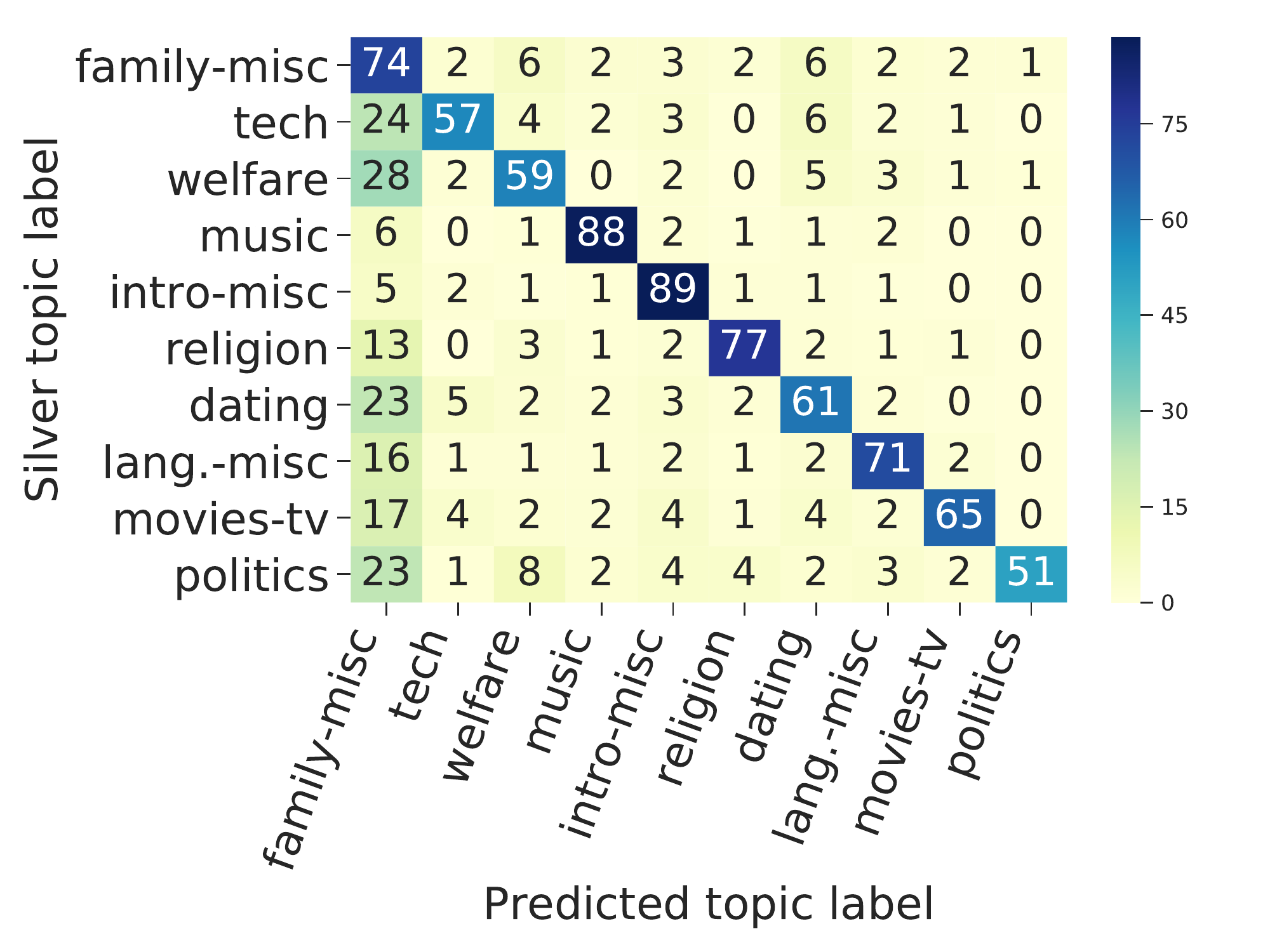}
  \vspace*{-5pt}
\caption{Confusion matrix for ST model trained on 20 hours of Spanish-English data. Each cell represents the percentage of the silver topic labels predicted as the $x$-axis label, with each row summing to 100\%.}
\label{fig:eval_topics_conf_matrix}
\end{figure}

Since topics vary in frequency, we look at label-specific accuracy to see if the ST models are simply predicting frequent topics correctly.
Figure~\ref{fig:eval_topics_conf_matrix} shows a normalized confusion matrix for the {\em ST-20h} model. Each row sums to 100\%, representing the distribution of predicted topics for any given silver topic, so the numbers on the diagonal can be interpreted as the topic-wise recall. For example, a prediction of {\em music} recalls  88\% of the relevant speech segments. We see that the model has a recall of more than 50\% for all 10 topics, making it quite effective for our motivating task.
The {\em family-misc} topic (capturing small-talk) is often predicted when other silver topics are present, 
with, for instance, 23\% of the silver {\em dating} topics predicted as {\em family-misc}.

\section{Related work}

We have shown that %even
low-quality ST can be useful for speech classification.
Previous work has also looked at speech analysis without high-quality ASR. In a task quite related to ours, \cite{dredze2010nlp} showed how to cluster speech segments in a completely unsupervised way. In contrast, we learn to classify speech using supervision, but what is important about our result is it shows that a small amount of supervision goes a long way.

A slightly different approach to quickly analyse speech, is the established task of \emph{keyword spotting}, which asks whether any of a specific set of keywords appears in a segment~\cite{wilpon+etal_assp90,garcia+gish_icassp06}. Recent studies have extended the early work to end-to-end keyword spotting~\cite{Palaz2016JointlyLT,menon2018fast} and to semantic keyword retrieval, where non-exact but relevant keyword matches are  retrieved~\cite{chelba+etal_ieee08,li+etal_asru13,lee+etal_taslp15}. In all these studies, the query and search languages are the same, while we consider the cross-lingual~case.

There has been some limited work on cross-lingual keyword spotting.
\cite{sheridan+etal_sigir97} introduced a baseline system which combined ASR and text translation to build a German speech retrieval system using French text queries. But source language transcriptions to train ASR are unlikely to be available in our scenarios of interest.
Some recent studies have attempted to use vision as a complementary modality to do cross-lingual retrieval~\cite{kamper2018visually,harwath2018vision}.
However, to the best of our knowledge, cross-lingual topic classification for speech has not been considered elsewhere.

\section{Conclusions and future work}
Our results show that poor speech translation can still be useful for speech classification in low-resource settings. 
By varying the amount of training data, we found that ST systems trained on as little as 10 hours (around 8K parallel utterances) of Spanish-English data  produce translations which still allow topics to be correctly classified in 61\% of input speech segments, outperforming a majority baseline.
With 20 hours of parallel data, accuracy is more than 70\%.

Since this is the first work in cross-lingual topic classification, there are a number of interesting avenues for future work.
We used our ST model as an off-the-shelf system, and did not tune its performance for the topic prediction task. We hope future work will improve accuracy further.
We used silver labels to evaluate our approach---this allowed us to compare several different settings using an objective metric.
However, human annotations of topics will be the next step.
We also used a pipelined approach of ST followed by classification.
An alternative would be to train a topic classifier on input speech directly, but we speculate that this would require more substantial resources.
Cross-lingual topic modeling may also be useful when the target language is high-resource; %. Here,
we learned target topics just from the 20 hours of translations, but in future work, we could use a larger text corpus in the high-resource language to learn a more general topic model covering a wider set of topics, and/or combine it with keyword lists curated for specific scenarios like disaster recovery~\cite{Olteanu2014CrisisLexAL}.

\section{Acknowledgements}
This work was supported in part by a James S McDonnell Foundation Scholar Award for SG and a Google Faculty Research Award for HK. We thank Ida Szubert, Marco Damonte, and Clara Vania for helpful comments on drafts of this paper.

\vfill\pagebreak


\begingroup
\newpage
\ninept
\bibliographystyle{IEEEtran}
\bibliography{refs}
\endgroup

% \iffalse
\iftrue

% \newpage

% ~ 

\newpage 

\appendix

\section{Using NMF for topic modeling}
\label{sec:appendix_nmf}
We now describe how we learn topics using NMF. 
Given a set of text documents as input, the model will output (1) for each document, a distribution over the selected number of topics (henceforth, the {\em document-topic} distribution), and (2) for each topic, a distribution over the set of unique terms in the text (henceforth, the {\em topic-term} distribution).

\subsection{Text processing}
\label{ssub:topic_text_processing}
Our training set ({\em train20h}) has 1080 English sentences. We start by generating a {\em tf-idf} representation for each of these. The English text contains 170K tokens and 6K terms (vocabulary size). 
As we are looking for topics which are coarse-level categories, we do not use the entire vocabulary, but instead focus only on the high importance terms. We lowercase the English translations and remove all punctuation, and stopwords. We further remove the terms occurring in more than 10\% of the documents and those which occur in less than 2 documents, keeping only the 1000 most frequent out of the remaining.

After preprocessing the training set, we have a feature matrix $V$ with dimensions $1080\times1000$, where each row is a document, and each column represents the {\em tf-idf} scores over the 1000 selected terms. The feature matrix will be sparse as only a few terms would occur in a document, and will also be non-negative as {\em tf-idf} values are greater than or equal to 0.

\subsection{Learning topics}
\label{ssub:learning_topics}
NMF is a matrix factorization method, which given the matrix $V$, factorizes it into two matrices: $W$ with dimensions $1080\times t$ (long-narrow), and $H$ with dimensions $t\times1000$ (short-wide), where  $t$ is a hyper-parameter. Figure~\ref{fig:nmf} shows this decomposition when $t$ is set to $10$. 

$$V~\approx~W~\times~H$$

In the context of topic modeling, $t$ is the number of topics we want to learn; $W$ is the {\em document-topic} distribution, where for each document (row) the column with the highest value is the most-likely topic; and $H$ is the {\em topic-term} distribution, where each row is a topic, and the columns with the highest values are terms most relevant to it.

\begin{figure}[t]
  \centering
  \includegraphics[width=0.9\linewidth]{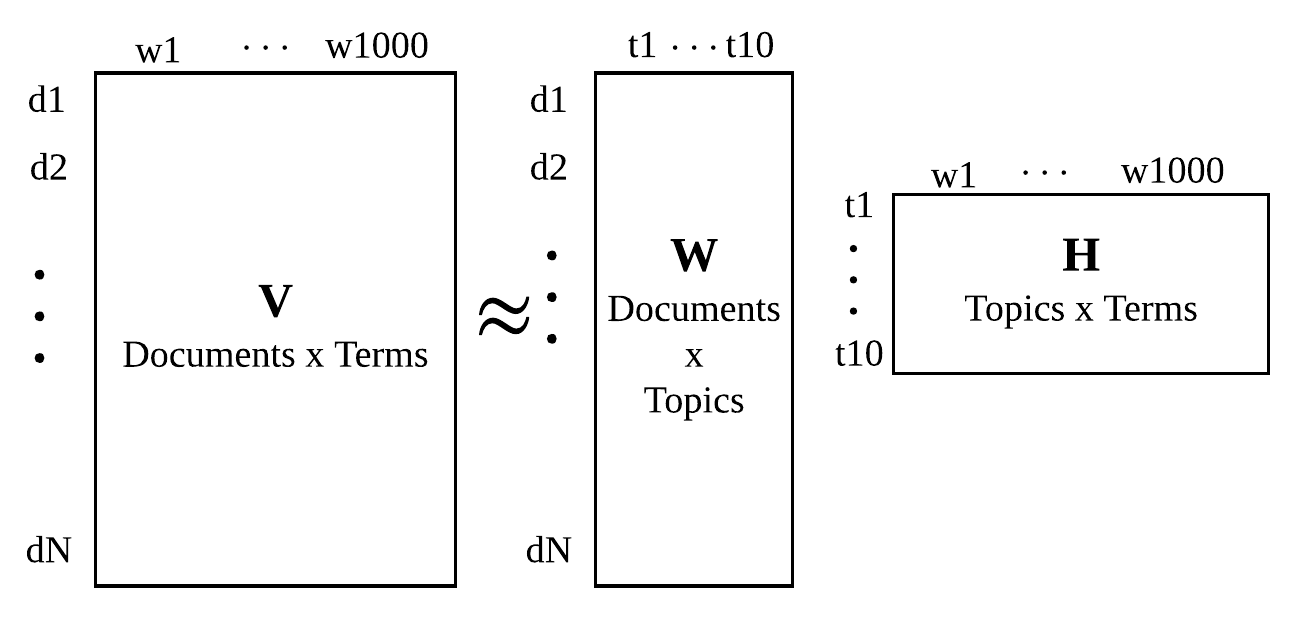}
  \caption{Nonnegative Matrix Factorization. $V$ is the document-term matrix, where $d$ is each document; $N$ is the number of documents; $w1$ to $w1000$ are the terms selected as features; and $t1$ to $t10$ are the topics.}
  \label{fig:nmf}
\end{figure}

The values for $W$ and $H$ are numerically approximated using a multiplicative update rule~\cite{lee2001algorithms}, with the Frobenius norm of the reconstruction error as the objective function. In this work, we use the machine-learning toolkit {\em scikit-learn}~\cite{scikitlearn} for feature extraction, and to perform NMF, using default values as described at scikit-learn.org.

\subsection{Making topic predictions}
\label{ssub:topic_predictions}
Using our {\em topic-term} distribution matrix $H$, we can now make topic predictions for new text input. Our evaluation set ({\em eval100h}) has 5376 English sentences. For each of these, we have the {\em gold} text, and also the ST model output. We preprocess and represent these using the same procedure as before (\ref{ssub:topic_text_processing}) giving us the feature matrix $V^{'}_{gold}$ for {\em gold}, and $V^{'}_{ST}$ for ST output, each with dimensions $5376\times1000$. Our goal is to learn the {\em document-topic} distributions $W^{'}_{gold}$ and $W^{'}_{ST}$, where:

$$V^{'}_{gold}~\approx~W^{'}_{gold}~\times~H$$
$$V^{'}_{ST}~\approx~W^{'}_{ST}~\times~H$$

The values for each $W^{'}$ matrix are again numerically approximated using the same objective function as before, but keeping $H$ fixed. 

\subsection{Silver labels and evaluation}
\label{ssub:evaluation}
We use the highest scoring topic for each document as the prediction. The {\em silver} labels are therefore computed as $argmax(W^{'}_{gold})$, and for ST as $argmax(W^{'}_{ST})$. We can now compute the accuracy over these two sets of predictions.

\section{Fisher corpus: assigned topics}% assigned to callers}
\label{sec:appendix_topics_assigned}
Figure~\ref{fig:topics_assigned} shows the topics assigned to callers in the Fisher speech corpus. Some topic prompts overlap, for example, {\em music-preference} asks callers to discuss what kind of music they like to listen to, and {\em music-social-message} asks them to discuss the social impact of music. For both these topics, we would expect the text to contain similar terms. Similarly the topics {\em cellphones-usage}, {\em tech-devices} and {\em telemarketing-spam} also overlap. Such differences might be difficult for an unsupervised topic modeling algorithm to pick up.

\begin{figure}[t]
  \centering
  \includegraphics[width=\linewidth]{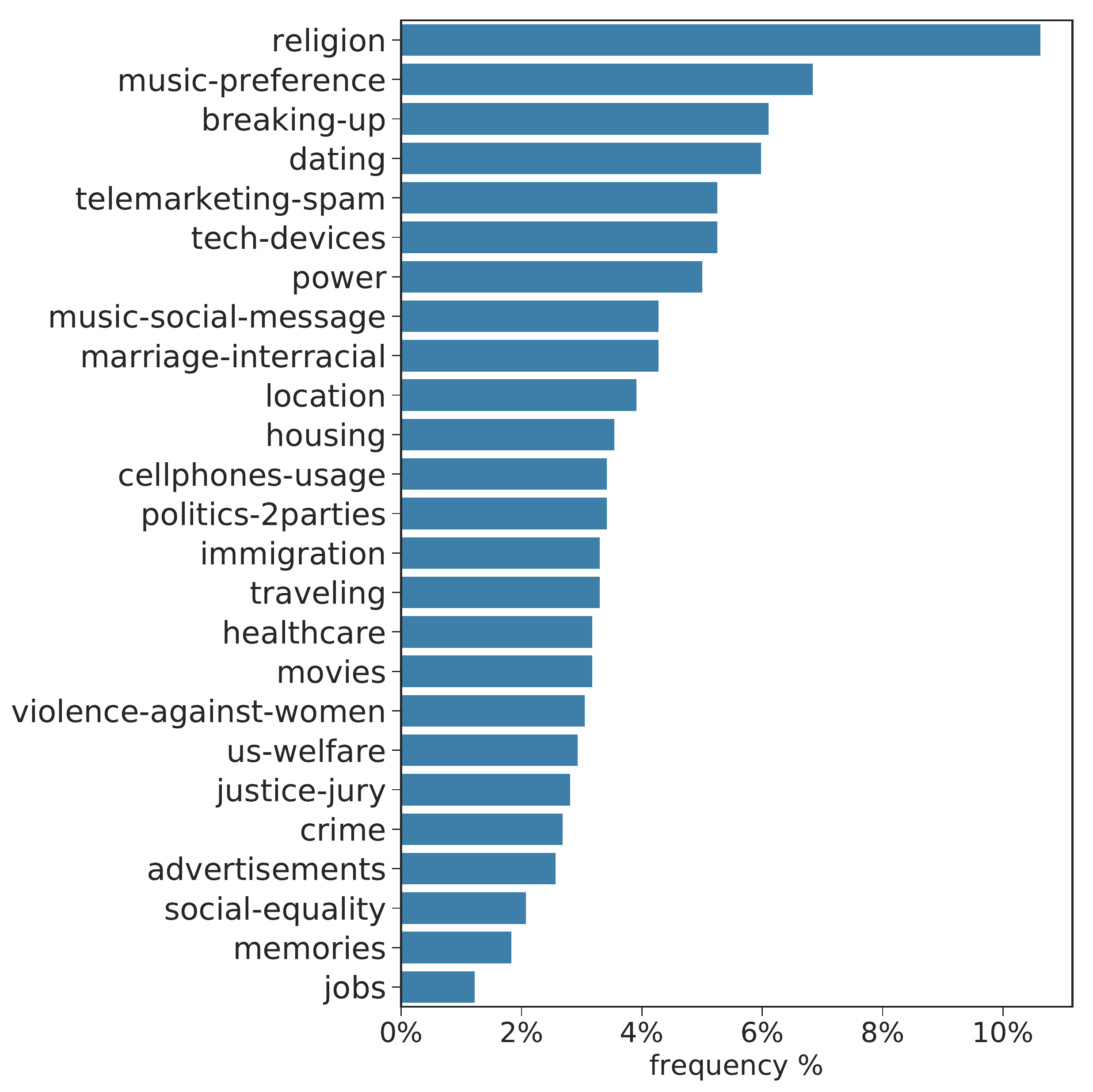}
  \caption{Topics assigned to callers in the Fisher dataset, as a percentage of the 819 calls.}
  \label{fig:topics_assigned}
\end{figure}

Table~\ref{tab:topics_n25_train160h} shows the topics learned by NMF by using human English translations from the entire 160 hours of training data as input, when the number of topics is set to 25.
We observe that some new topics are found that were not discovered by the 20hr/10-topic model and that match the assigned topic prompts, such as {\em juries} and {\em housing}. However, there are also several incoherent topics, and we don't find a major improvement over the topics learned by just using 20 hours of training data, with the number of topics set to 10.

\begin{table}[h]
\begin{center}
   \footnotesize
  \begin{tabularx}{\linewidth}{c@{\hspace{1mm}}l@{\hspace{2.5mm}}X}
    \toprule
    {\bf id} & {\bf Assigned name} & {\bf Most informative words} \\
    \midrule
    1 & --- & told, went, maybe, take, ll \\
    2 & music & music, listen, dance, play, classical \\
    3 & intro & hello, name, speaking, topic, talked \\
    4 & religion & religion, religions, catholic, church, religious \\
    5 & welfare & pay, insurance, expensive, doctor, health \\
    6 & languages & spanish, speak, english, language, learn \\
    7 & relationships & married, marriage, got, divorced, together \\
    8 & tech-marketing & phone, cell, telephone, calls, cellular \\
    9 & --- & hundred, dollars, thousand, five, fifty \\
    10 & chatter & cold, snow, winter, hot, weather \\
    11 & --- & puerto, rico, rican, born, ricans \\
    12 & movies-tv & watch, movies, movie, tv, kids \\
    13 & --- & city, mexico, big, lived, living \\
    14 & --- & huh, gonna, give, us, lets \\
    15 & --- & yea, tv, lots, pretty, expensive \\
    16 & locations & york, manhattan, bronx, carolina, panama \\
    17 & internet-dating & internet, computer, use, met, information \\
    18 & --- & old, twenty, kids, thirty, five \\
    19 & politics & power, countries, world, government, help \\
    20 & housing & house, buy, rent, apartment, houses \\
    21 & juries & system, jury, health, social, help \\
    22 & religion & god, believe, church, bible, thank \\
    23 & violence & women, man, woman, men, abuse \\
    24 & intro & hi, fine, name, philadelphia, evening \\
    25 & welfare & money, give, make, help, need \\
  \bottomrule
  \end{tabularx}
  \end{center}
  \caption{Topics discovered using human translated text from the full 160hr Fisher training set. We set the number of topics to 25. We assign the topic names manually, and use {\em ---} where the topic clustering is not very clear.
  }
%   \vspace{-1em}}
  \label{tab:topics_n25_train160h}
\end{table}

\section{Tracking topic drift over conversations}
\label{sec:appendix_topics_drift}
To measure how often speakers stray from assigned topic prompts,  we take a closer look at the calls in {\em train20h} with the assigned prompt of {\em religion}. This is the most frequently assigned prompt in the Fisher dataset (17 calls in {\em train20h}). We also select this topic for further analysis as it contains terms which are strongly indicative, such as {\em god}, {\em bible}, etc. and should be relatively easier for our topic model to detect. 

Figure~\ref{fig:religion_topic_drift} shows the trend of discussion topics over time. Overall, only 36\% of the total dialog segments in these calls have the silver label {\em religion}, and the most frequently assigned label is {\em family-misc} with 46\%. We observe that the first segment is often labeled as {\em intro-misc}, around 70\% of the time, which is expected as speakers begin by introducing themselves. Figure~\ref{fig:music_topic_drift} shows that a similar trend emerges for calls assigned the prompt {\em music} (14 calls in {\em train20h}). Silver labels for {\em music} account for 45\% of the call segments and {\em family-misc} for around 38\%.

\begin{figure}[t]
\centering

\includegraphics[width=.45\textwidth]{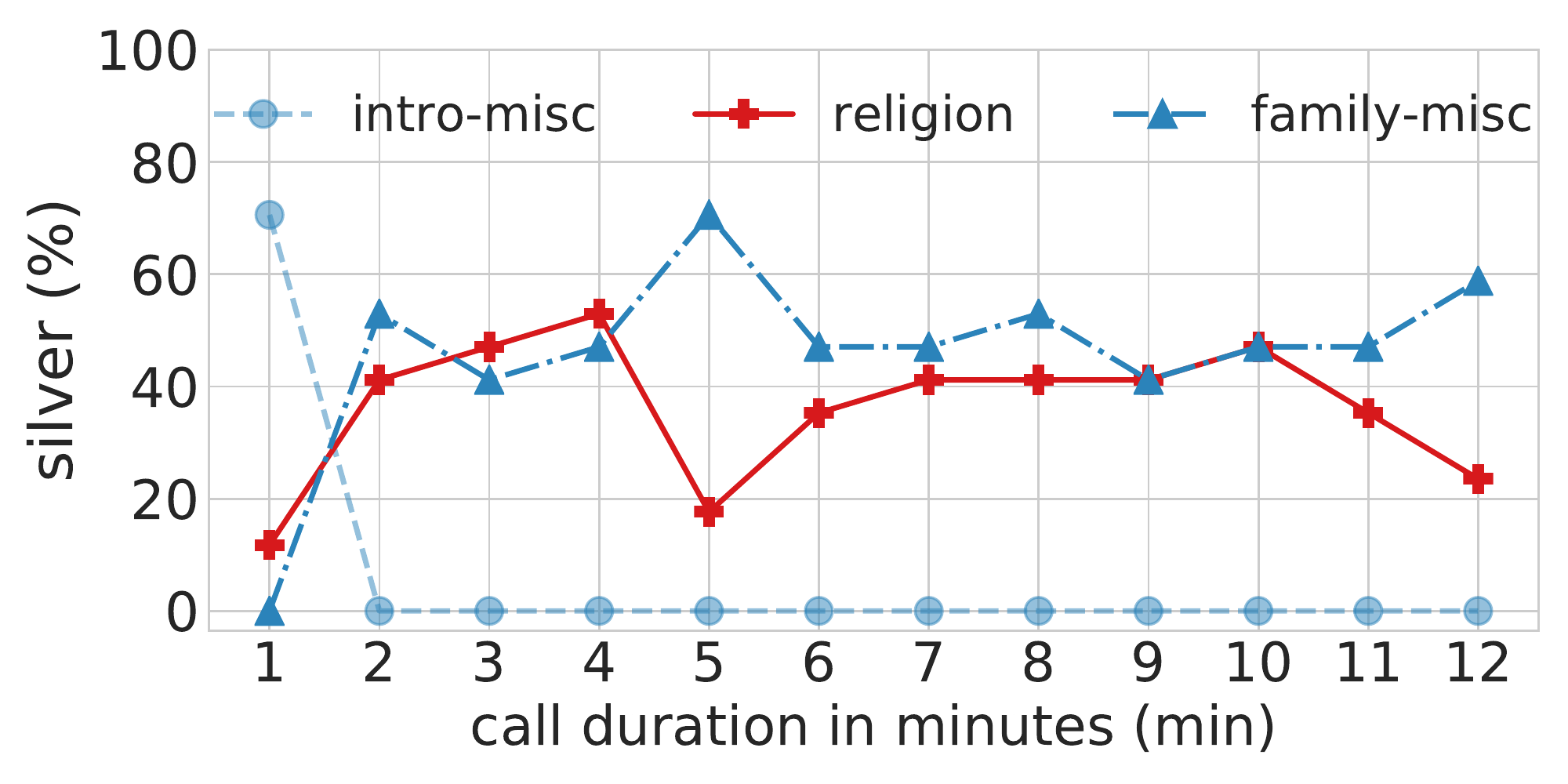}
  \vspace*{-5pt}
\caption{Tracking silver labels over time for calls where the assigned prompt is {\em religion}. Total of 17 calls in {\em train20h}.}
\label{fig:religion_topic_drift}

\includegraphics[width=.45\textwidth]{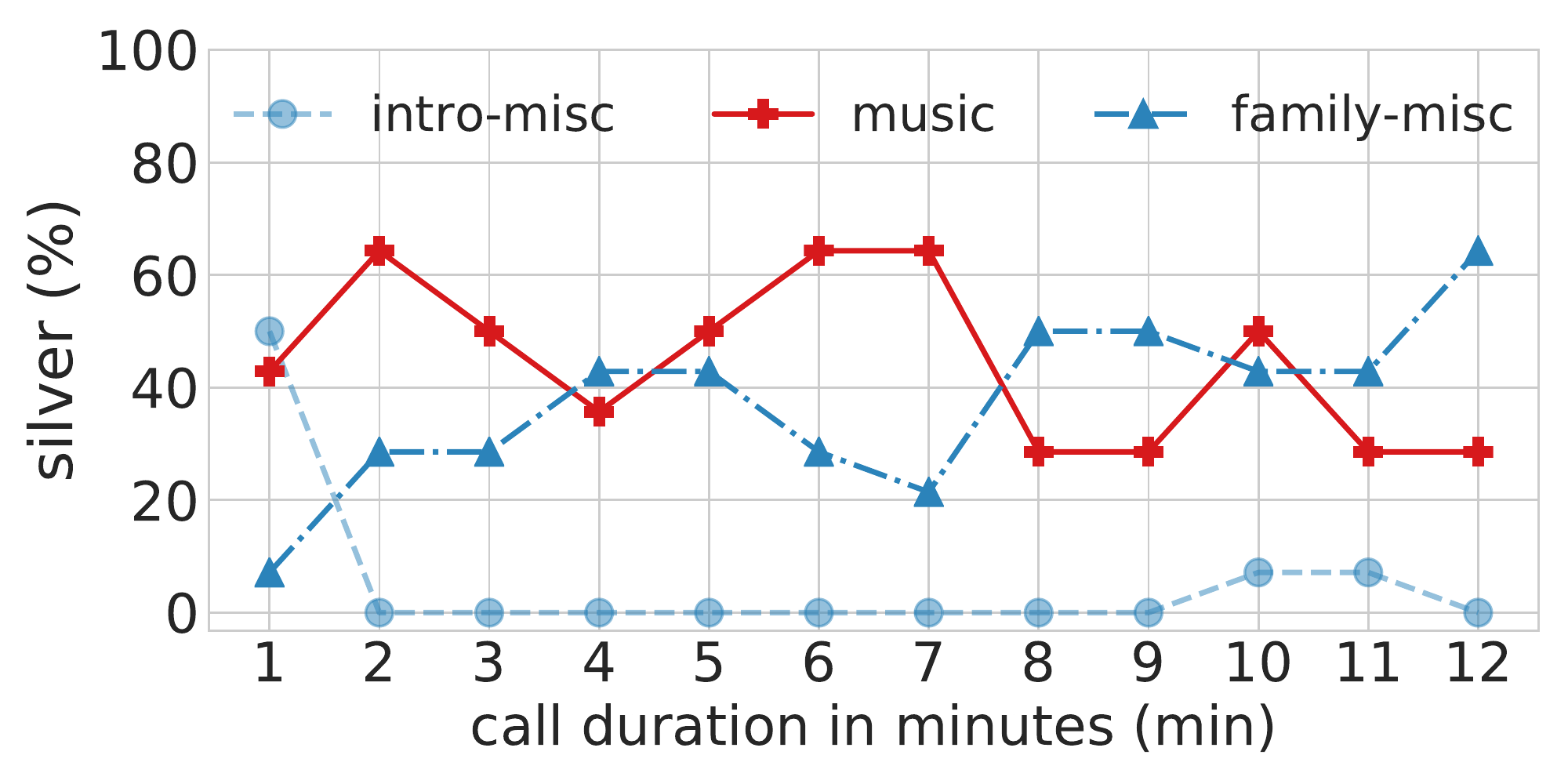}
  \vspace*{-5pt}

\caption{Tracking silver labels over time for calls where the assigned prompt is {\em music}. Total of 14 calls in {\em train20h}.}
\label{fig:music_topic_drift}
\end{figure}

\fi
\end{document}